\documentclass{article}

\usepackage[english]{babel}

\usepackage[letterpaper,top=2cm,bottom=2cm,left=2.5cm,right=2.5cm,marginparwidth=1.75cm]{geometry}

\usepackage{amsmath}
\usepackage{graphicx}
\usepackage[colorlinks=true, allcolors=blue]{hyperref}
\usepackage{authblk}
\usepackage{hyperref}
\usepackage{makecell}
\usepackage{booktabs}
\usepackage{tabularx}
\usepackage{xurl}
\usepackage{float}
\usepackage[
  backend=biber,
  style=nature, 
  sorting=none,
  url=false,      
  doi=false,
]{biblatex}
\usepackage[title]{appendix}
\AtBeginBibliography{%
  \DeclareFieldFormat[online]{title}{#1}%
  \DeclareFieldFormat[inproceedings]{title}{#1}%
  \DeclareFieldFormat[misc]{title}{#1}%
}
\AtEveryBibitem{%
  \clearfield{issn}%
  \clearfield{isbn}%
}
\DefineBibliographyStrings{english}{in = {}}
\usepackage{listings}
\title{\textit{PARC}: An Autonomous Self-Reflective Coding Agent for Robust Execution of Long-Horizon Tasks}
\author{Yuki~Orimo}
\author{Iori~Kurata\thanks{Equal contributions}}
\author{Hodaka~Mori\protect\footnotemark[1]}
\author{Ryuhei~Okuno\protect\footnotemark[1]}
\author{Ryohto~Sawada\protect\footnotemark[1]}
\author{Daisuke~Okanohara}

\affil{Preferred Networks, Inc.}

\date{}

\begin{document}
\maketitle

\begin{abstract}
We introduce \textit{PARC}, a coding agent for the autonomous and robust execution of long-horizon computational tasks.  
\textit{PARC} is built on a hierarchical multi-agent architecture incorporating task planning, execution, and a mechanism that evaluates its own actions and their outcomes from an independent context and provides feedback, namely self-assessment and self-feedback. This design enables \textit{PARC} to detect and correct high-level strategic errors and sustain progress without human intervention. 
We evaluate \textit{PARC} across computational science and data science tasks. In materials science, it autonomously reproduces key results from studies on lithium-ion conduction and alloy segregation. In particular, it coordinates dozens of parallel simulation tasks, each requiring roughly 43 hours of computation, managing orchestration, monitoring, and error correction end-to-end. In Kaggle-based experiments, starting from minimal natural-language instructions, \textit{PARC} conducts data analysis and implements search strategies, producing solutions competitive with human-engineered baselines. 
These results highlight the potential of integrating a hierarchical multi-agent system with self-assessment and self-feedback to enable AI systems capable of independent, large-scale scientific and analytical work.
\end{abstract}

\section{Introduction}
Engaging AI agents in long-horizon tasks has long been a central goal in artificial intelligence (AI) research. In this context, tasks that can be completed entirely within a command-line interface represent a domain that is particularly amenable to the application of AI agents. Indeed, coding agents such as Cline\cite{cline}, Claude Code\cite{claude-code}, and Codex\cite{openai-codex-cli} have made remarkable strides in recent years. These agents have become indispensable tools for software engineers, capable of handling a wide variety of coding tasks. However, the execution of complex tasks requiring numerous steps or extended periods of time, namely "long-horizon tasks", remains a persistent challenge. One concrete illustration of this difficulty is SWE-Bench Pro~\cite{deng2025swebenchpro}, a benchmark that evaluates coding agents on complex, long-horizon software engineering tasks drawn from real enterprise codebases. On the SWE-Bench Pro (Public Dataset), current best LLMs reach only around 40\% resolve rate, and performance on the more challenging SWE-Bench Pro (Commercial Dataset) is expected to remain even lower.

We believe that the challenges associated with these long-horizon tasks are not solely attributed to insufficient performance of the underlying Large Language Models (LLMs). In fact, frontier LLMs of 2025, such as GPT-5, Claude Opus/Sonnet 4.5, and Gemini 3, have already achieved scores surpassing human experts on various benchmarks measuring coding capabilities, mathematical reasoning, and scientific knowledge~\cite{zhong2024o1,luo2025brainbench,brodeur2024physician,rein2023gpqa}. This observation suggests that the difficulty of long-horizon tasks arises not only from the underlying LLM capabilities but also from limitations of current agent designs. We therefore explicitly separate these two factors and show how far long-horizon performance can be pushed purely by improving the agent design given frontier LLMs.

Current mainstream coding agents, referred to as standard coding agents in this paper, employ a design that sequentially stacks all events into the context. These events include user interactions, reasoning and responses, tool calls, and tool execution results, which serve as input for determining the next action. Such a design is highly effective in short-term tasks where all relevant information can be stored within a single context window. However, in long-horizon tasks requiring numerous steps or extended periods to complete, this architecture can become a severe bottleneck regarding context saturation, error accumulation, and maintaining long-term perspective. In systems that proceed in a simple sequential manner, the probability of overall task success decreases exponentially as steps form a chain. Even with a per-step success rate as high as 99\%, a sequence of 100 steps has only about a 37\% chance of all succeeding. Moreover, because of the sequential context design, while standard coding agents can handle local and short-term failures caused by immediate actions, they find it extremely difficult to recognize and correct approach-level failures stemming from more fundamental issues such as an inappropriate previously adopted strategy or incorrect initial hypotheses and premises. For example, in a development task in scientific research involving hundreds of steps, an agent may make a single mistaken choice of approach along the way without realizing it and eventually run an incorrect simulation and analyze its erroneous results.

While these challenges might be overcome by further advancements in LLMs, we believe they can also be addressed by a more powerful agent harness or architecture. For example, AlphaEvolve~\cite{novikov2025alphaevolve}, although in a context different from that of pure coding agents, models the process of discovery and provides a harness that allows the LLM to iteratively refine solutions, which enables advanced algorithmic discovery and problem-solving involving long-term trial and error. These are almost impossible for standard coding agents alone.

In this work, we present \textit{PARC}, an agent capable of autonomously executing complex, long-horizon tasks. 
The name is derived from ``\textit{Preferred}\footnote{``\textit{Preferred}" derives from our company name.} Autonomous self-Reflective Coding agent". \textit{PARC} uses the standard coding agent as a fundamental component while incorporating self-assessment and self-feedback mechanisms into its architecture, inspired by approaches using LLMs to evaluate or improve generated outputs, such as LLM-as-a-Judge~\cite{zheng2023judging} and Self-Refine~\cite{madaan2023selfrefine}. This enables the agent to self-reflect on its own actions and their outcomes from a holistic and long-term perspective, allowing it to perform approach-level corrections beyond merely handling local failures. The improvements we report on long-horizon tasks thus stem from the \textit{PARC} agent architecture rather than from changes to the underlying LLMs.

To demonstrate the general problem-solving capabilities of \textit{PARC}, we evaluated its performance across multiple domains. In the context of AI for Science, we conducted demonstration experiments using challenges in materials science. In these experiments, \textit{PARC} successfully and autonomously completed multiple simulation research tasks aimed at partially reproducing results reported in published studies. Furthermore, in experiments using recent Kaggle competitions covering both machine learning and algorithmic implementation challenges, \textit{PARC} succeeded in generating solutions comparable to those created by humans when provided with only the problem statement and minimal instructions.

In summary, this paper makes the following contributions:
\begin{itemize}
\item We design \textit{PARC}, an autonomous coding agent with self-assessment and self-feedback mechanisms that enables stable and reliable execution of complex long-horizon tasks.

\item We demonstrate that this agent can autonomously execute complex multi-stage workflows on the order of a few tens of tasks and roughly a hundred steps in domains such as scientific research and data science, including nontrivial simulation studies and challenging predictive modeling and algorithmic implementation problems.
\end{itemize}

In the remainder of this paper, we first outline the design of \textit{PARC}. Subsequently, through representative case studies, we demonstrate its autonomous execution capabilities for long-horizon tasks and its problem-solving process.

\section{Approach}

\textit{PARC} is a coding agent designed to autonomously and reliably execute complex long-horizon tasks. The architecture incorporates the concept of self-reflection mentioned in the introduction. This enables the agent to handle processes such as planning, execution, and self-assessment (self-reflection) in a structured manner.

\begin{figure}[H]
\centering
\includegraphics[width=0.95\linewidth]{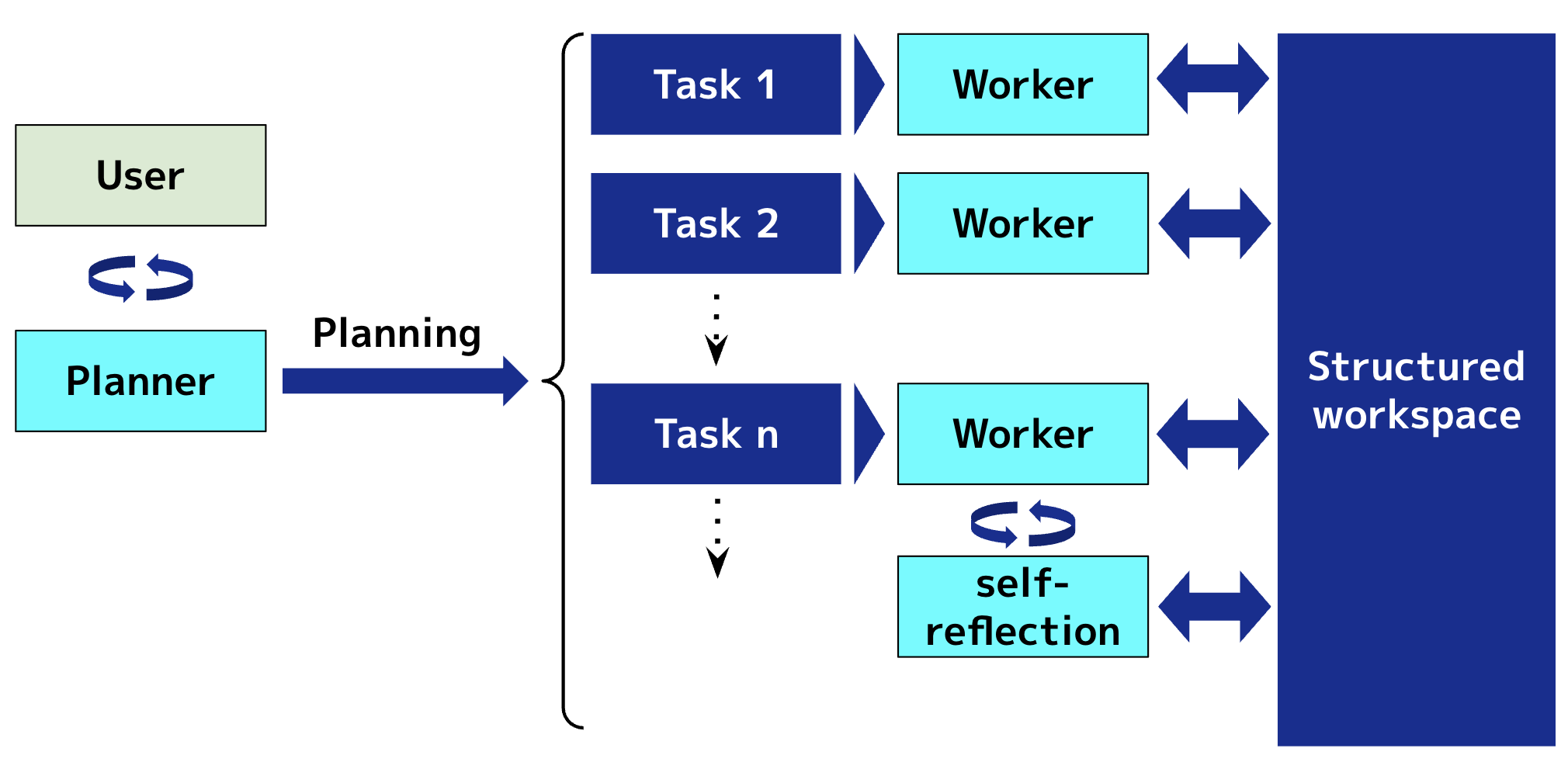}
\caption{\label{fig:simple_workflow}
Schematic overview of the \textit{PARC} workflow. A user interacts with the planner to perform planning and generate a sequence of tasks. Workers then execute these tasks sequentially, writing task results to and reading preceding results from the structured workspace. Through self-reflection (self-assessment and self-feedback), the workers can robustly make progress on long-horizon tasks. 
}
\end{figure}

\textit{PARC} operates as a multi-agent system, where a \textit{planner} formulates plans and multiple \textit{workers} execute them, following a plan-and-execute pattern as shown in Fig.~\ref{fig:simple_workflow}. First, the planner interacts with the user and constructs a plan for the project and a sequence of tasks required to carry it out. Each task is executed by a worker that operates in an independent context whose scope is limited to that task. This design is effective in avoiding the context window limitations and contamination issues that standard coding agents often face. The planner and workers are built on top of a standard coding agent and inherit its task execution capabilities. In addition, in order to execute long-horizon tasks in a robust manner, \textit{PARC} incorporates a self-feedback system based on self-assessment. This self-feedback covers not only local errors, for example failures during code execution, but also the consistency of the resulting sequence of work and the validity of the approach adopted to accomplish the task. As a result, even if errors occur while executing complex long tasks, the agent can autonomously correct them and reliably produce correct outcomes.

In the planning phase, the planner interacts with the user to construct a project plan and a task sequence, that is, a list of tasks to be executed. Each task is designed so that an underlying worker can complete it on its own or at least come close to doing so. The resulting plan and task sequence are then reviewed and approved by a human. After this approval, the agent proceeds to the next task execution phase. For this reason, the agent mainly targets problems for which it is reasonably possible to anticipate and decompose the set of tasks at the beginning of the project.

Once the plan is approved, the agent enters an autonomous task execution phase. Each worker advances its respective task step by step while using self-feedback. Self-feedback is also used to control the overall progress of the project. Specifically, when deciding whether to move on to the next task, the agent refers to this feedback and proceeds to the next task only if it judges that the outcome of the current task has reached a sufficient level of quality.

If the agent determines that there is a problem that cannot be resolved by further corrections at the level of individual tasks, it halts the project. Typical situations include cases where the project cannot proceed because earlier tasks fail to produce the results assumed in the plan and later tasks therefore lack the necessary prerequisites, and cases where, despite repeated corrections, the results of the tasks cannot be made to satisfy the requirements specified in the plan.
Through these mechanisms, \textit{PARC} ensures stability and reliability in long-horizon tasks. When a project is halted, the user can review the agent's outputs and, if necessary, revise the plan or provide new instructions so that the project can be resumed from the point at which it was stopped.

The agent performs its work on a structured workspace, that is, a working directory shared across all tasks. Files generated in a task, such as code, data, and configuration files, are stored in this workspace, and subsequent tasks can utilize them. For example, code developed in an earlier task is stored in a task-specific or shared directory of the workspace, making it straightforward for subsequent tasks to locate and execute it. In addition to this file-based sharing of information, the results and insights of individual tasks are aggregated and passed on to related subsequent tasks. This summary records, for example, where the outputs of each task are stored, what outcome each task reached, how to use code artifacts, and in what format any data artifacts are provided, so that subsequent tasks can make effective and smooth use of the results of earlier tasks. With this design, the workers for later tasks can efficiently inherit the context required to perform their tasks without directly accessing the detailed artifacts of earlier tasks such as large volumes of logs or numerous intermediate files.

This agent is designed to handle workflows where each unit of work is a task that a standard coding agent can perform or a task of comparable complexity. In such workflows, tasks may have dependency relationships. For example, in a computational science research project, one task may evaluate parameters to identify those that yield the best convergence in a simulation, and a subsequent task may use these results to perform a production run. 
As another example, in a data science project, one task may compare several models and hyperparameter settings using training data, and a subsequent task may use the chosen configuration to train the final model and produce predictions on a separate test dataset. These are only illustrative examples, and the agent can handle more complex cases as well.

\section{Demonstration}
In this section, we report the results of a series of experiments conducted to demonstrate the effectiveness of the proposed \textit{PARC}. We show that the self-assessment and self-feedback mechanism introduced in \textit{PARC} enables the autonomous execution of long-horizon tasks with high reliability, addressing challenges that would be difficult for conventional standard coding agents.
To ensure the reliability of the reported results, the generated solutions and task results were reviewed and validated by domain experts. Although \textit{PARC} supports multiple LLM providers and local models, all experiments were conducted using Claude Sonnet 4.5, which served as the best performing model in these tasks. As will be described in later subsections, the projects used for evaluation consisted of 10 - 20 tasks and required on the order of one to several days of autonomous execution in total, including program runtime. Since each task had 5--10 subtasks, each project had a complexity of roughly 100 steps on average. This provides an example of the problem scale that the architecture can solve, and the number of tasks or the overall execution time is not inherently limited to this scale by the architectural principles.

\subsection{Case Study: Lithium Diffusion in Solid Electrolytes}
\label{sec:lithium}

In this case study, we applied \textit{PARC} to a simulation task concerning lithium diffusion in a solid electrolyte with a complex crystal structure $\mathrm{Li_{10}GeP_2S_{11.5}O_{0.5}}$ (LGPS), with the aim of partially replicating the simulation study conducted in Ref.~\cite{sawada2024high}. LGPS, a sulfide solid electrolyte, is regarded as a key candidate for all-solid-state lithium batteries owing to its high lithium-ion conductivity at room temperature.
This computational science task involves determining the most stable structure via random search and structural optimization based on specified atomic coordinates and occupancy constraints, subsequently performing molecular dynamics (MD) simulations at multiple temperatures using the optimized structure, and finally calculating the activation energy for ion diffusion. The activation energy is derived from the Arrhenius plot of diffusion coefficients, which are calculated from the mean squared displacement (MSD) of lithium ions at each temperature.

\begin{figure}[H]
\centering
\includegraphics[width=0.95\linewidth]{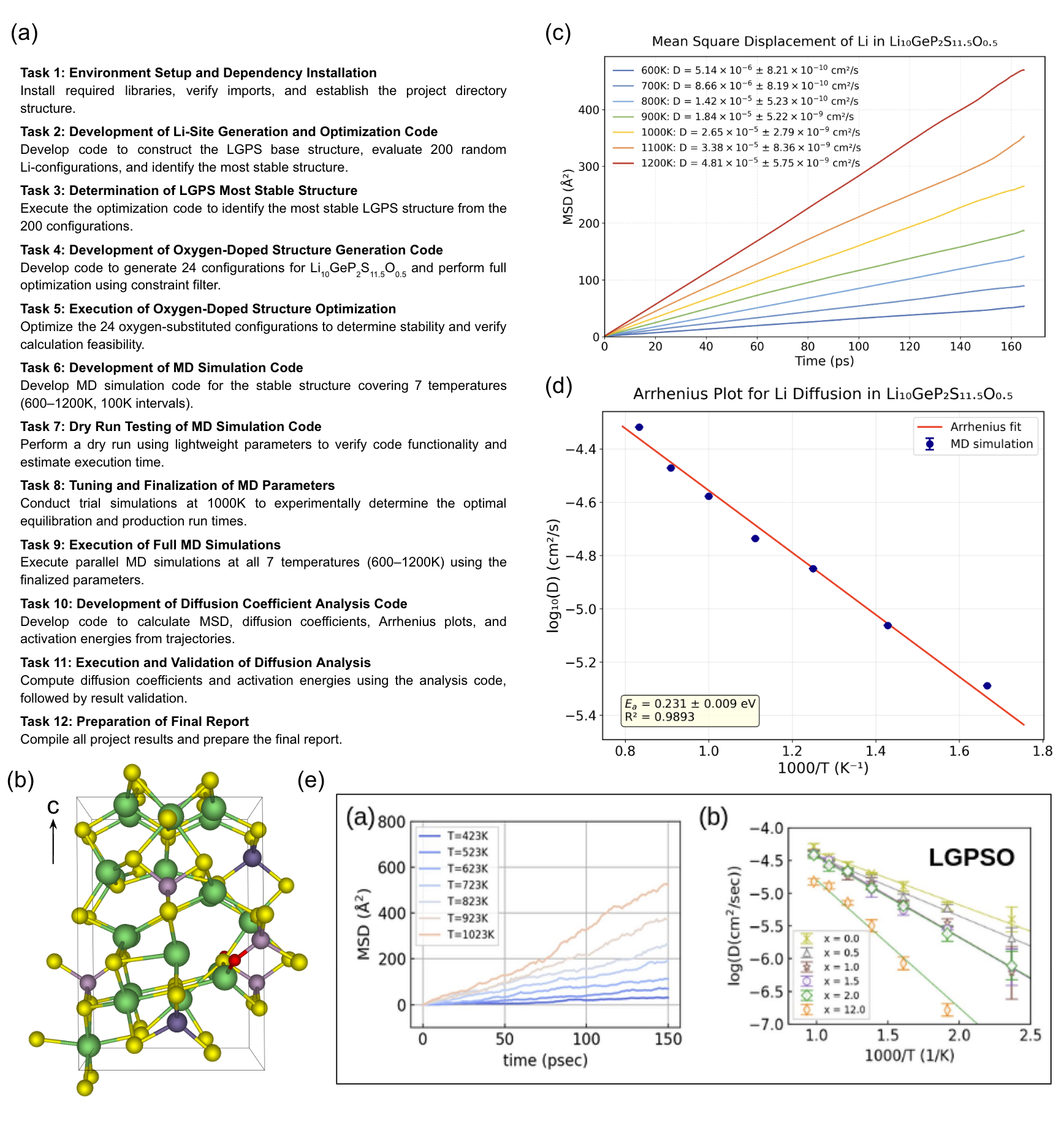}
\caption{\label{fig:sawada_paper}
Key outputs of \textit{PARC} on MD simulations of Li ion diffusion in the solid electrolyte $\mathrm{Li_{10}GeP_2S_{11.5}O_{0.5}}$.
(a) Overview of the task sequence generated by the planner.
(b) $\mathrm{Li_{10}GeP_2S_{11.5}O_{0.5}}$ structure generated by \textit{PARC}. This structure was visualized using VESTA~\cite{vesta}. Although the structure differs slightly from that in the original paper (Ref.~\cite{sawada2024high}) because the generation method involves random search, it was constructed using the correct procedure.
(c) Time evolution of MSD of Li ions calculated by \textit{PARC} from simulation results at each temperature. Although simulations were performed for 500 ps, the data covers approximately 165 ps because the trajectory was divided into three segments for block averaging.
(d) Arrhenius plot of diffusion coefficients calculated by \textit{PARC} from (c). The activation energy derived from the slope is 0.231 eV.
(e) Corresponding results reproduced from Figure 3(a) and (b) of the original paper (CC BY 4.0). Subpanel (a) displays the MSD of Li ions in the LGPS structure without O. Note that our results in Panel (c) plot the MSD in $\mathrm{Li_{10}GeP_2S_{11.5}O_{0.5}}$. Subpanel (b) displays Arrhenius plots for LGPSO structures; our results in Panel (d) correspond to the $x=0.5$ case in this graph.
}
\end{figure}

The specific instructions provided to \textit{PARC} are detailed in Appendix~\ref{appx:lithium}. To ensure results comparable to the original study, these instructions specified the physical system (material compositions and crystal structure), the quantities of interest (diffusion coefficients and activation energy), and a high-level workflow (random search for a low-energy structure, construction of oxygen-substituted models, and analysis via MSD and Arrhenius plots). By contrast, they did not prescribe concrete simulation algorithms or numerical settings; choices such as supercell size, MD protocol, temperature schedule, simulation duration, time step, ensemble, and convergence criteria were left open and were instead determined autonomously by \textit{PARC} during planning and execution. Upon receiving the instructions, \textit{PARC}'s planner investigated the necessary external libraries and tools and subsequently generated a task sequence as shown in Fig.~\ref{fig:sawada_paper}(a). Although the planner is capable of interacting with the user, in this instance, it formulated the plan without requesting additional instructions or asking questions. The task sequence outlines a comprehensive research workflow: starting with environment setup, it proceeds to the creation of LGPS and oxygen-doped structures, code development, parameter tuning through trials, execution of multiple simulations, and the verification and analysis of results, finally concluding with the compilation of a research report.

Through the execution of this task, \textit{PARC} obtained the results shown in Fig.~\ref{fig:sawada_paper}(b), (c), and (d), finally calculating an activation energy of 0.23 eV, which was the primary objective of the calculation. This value is sufficiently close to the approximately 0.18 eV reported in the original study~\cite{sawada2024high} (captured from Subpanel~(b) in Fig.~\ref{fig:sawada_paper}(e)). The lack of exact agreement is likely mainly attributed to differences in the LGPS structures used in the simulations. The original study used a random search to generate the LGPS structure; since we instructed \textit{PARC} to employ the same method, the resulting structure has randomness in the positions of Li atoms. In addition to this structural difference, the activation energy extracted from MD trajectories inevitably carries statistical uncertainty due to the finite simulation time and stochastic diffusion pathways, and can also be affected by minor differences in analysis procedures and simulation parameters (e.g., simulation time, cell size) that were not specified in the instructions. Furthermore, in the original study, the percolation transition induced by oxygen doping changes the activation energy by about 0.11 eV, which provides a reference for the characteristic energy scale of diffusion-pathway changes in this material. Therefore, the 0.05 eV discrepancy between 0.23 eV and 0.18 eV is small compared with this scale. Despite these discrepancies, as far as we could verify, no significant bugs or errors were found in the programs and the simulations.

The self-feedback mechanism proved effective across the entire task sequence, ranging from code bug detection and simulation flow verification to confirming the stability of generated structures and judging the correctness of simulation results. Without this self-feedback, the task would almost certainly have failed midway or produced incorrect simulation results. In particular, during the verification of the analysis results in Task 11, prompted by self-feedback on the irregular temperature dependence, accuracy of the Arrhenius fit, and limited statistical sampling, the agent questioned the statistical reliability of the simulations conducted under conditions of relatively low computational cost and, by redoing the work from Task 8 under stricter conditions, ultimately obtained a more accurate result.

\subsection{Case Study: Effect of Light Interstitials on Cr-Ni Alloy}
\label{sec:alloy}

In this case study, we applied \textit{PARC} to the task of partially reproducing a study \cite{DOLEZAL2025121221} that evaluated the segregation behavior of light interstitial elements in a $\mathrm{Cr_{30}Ni}$ alloy and the associated structural changes in the Cr-Ni matrix using Monte Carlo (MC) simulation. Ni-based superalloys are essential for high-temperature components, where trace additions of light interstitials like B and N strongly affect grain-boundary strength. Clarifying these atomistic segregation mechanisms is thus important for alloy design. In this context, the task requires the agent to implement and execute a Monte Carlo simulation based on specified complex transition rules and to quantitatively analyze the resulting segregation behavior and structural changes in the matrix.

We instructed \textit{PARC} to implement the MC simulation and perform five runs for each of the seven systems ($\mathrm{Cr_{30}Ni}$ alloy and systems doped with 1, 4, and 10 at.\% B or N). The simulations were conducted using a $6\times6\times6$ supercell at 1073 K, employing five specified trial moves (Swap, Relocate, etc.). Additionally, we requested the agent to generate a final output plotting the changes in structural fractions (FCC, HCP, BCC, other) against the interstitial element concentration, utilizing the polyhedral template matching analysis. The actual instructions provided are shown in Appendix \ref{appx:alloy}.

\begin{figure}[H]
\centering
\includegraphics[width=0.95\linewidth]{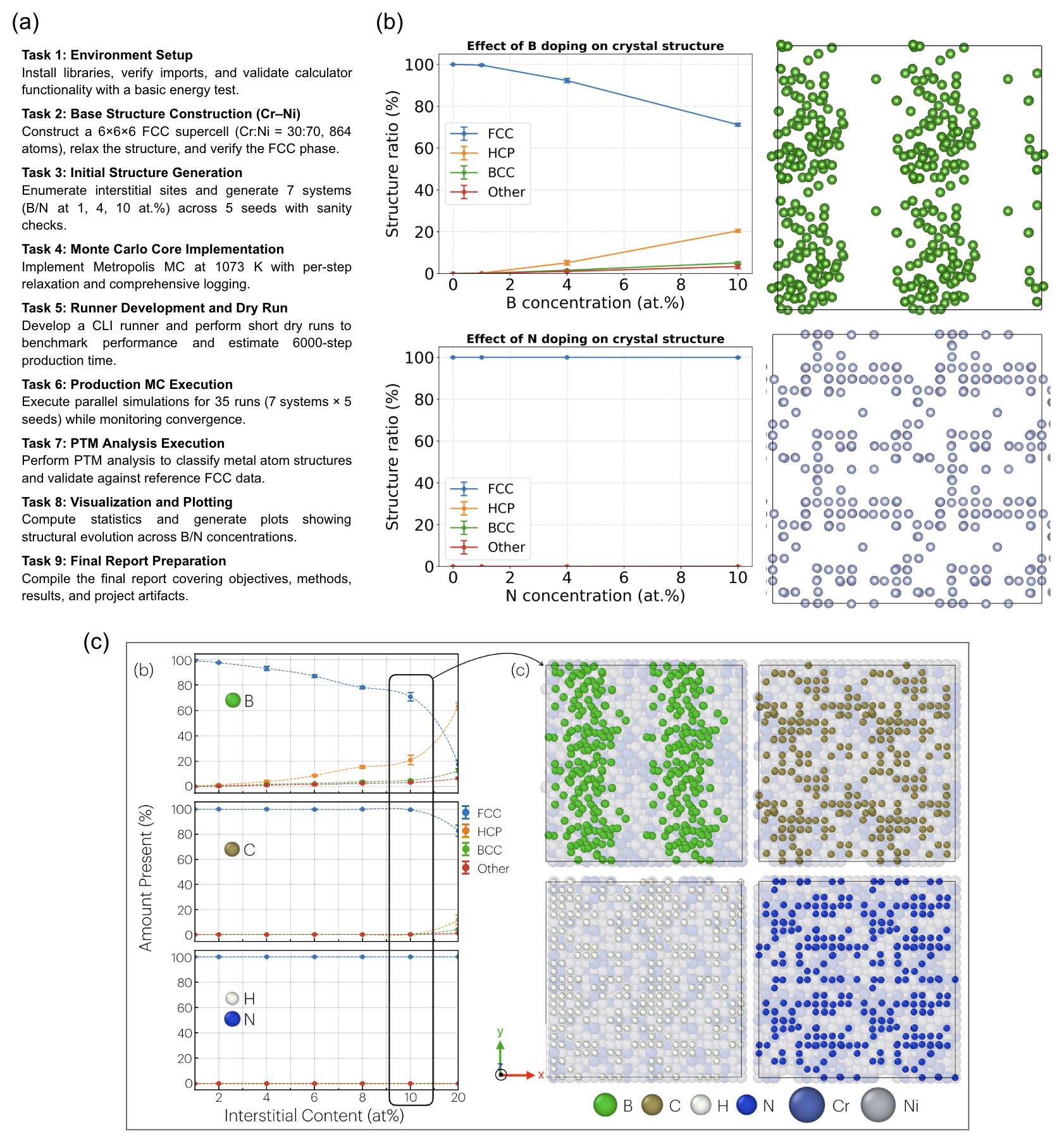}
\caption{\label{fig:juli_paper}
Key outputs of \textit{PARC} on simulations of the effect of light interstitials in $\mathrm{Cr_{30}Ni}$ alloys.
(a) Overview of the task sequence generated by the planner.
(b) Simulation results of crystal structure stability regarding interstitial element species by \textit{PARC}. Structures were visualized using VESTA~\cite{vesta}.
Top: Evolution of structural fractions (FCC/HCP/BCC/Other) with B doping (1, 4, and 10 at.\%) and the final atomic configuration at 10 at.\% (only B atoms are visualized). B doping reduces the FCC fraction with increasing concentration and leads to segregation.
Bottom: Evolution of structural fractions with N doping (1, 4, and 10 at.\%) and the final configuration at 10 at.\% (only N atoms are visualized). N doping maintains the FCC structure up to 10 at.\%. 
(c) Corresponding results from the original study~\cite{DOLEZAL2025121221}. Our results are consistent with the reference, demonstrating that \textit{PARC} correctly executed the implementation, simulation, and analysis. 
Panel (c) is reproduced from Figure 2 (b–c) of the preprint \cite{Dole_al_2025}, with permission from the authors.
}
\end{figure}

In response to these instructions, the planner generated a task sequence with 9 steps (Fig. \ref{fig:juli_paper}(a)). This plan represents a comprehensive research workflow, starting with environment setup and proceeding through initial structure generation, MC simulator implementation, verification via dry runs, and parallel execution (35 simulations in total), concluding with analysis and report generation.

\textit{PARC} successfully completed all assigned tasks, yielding the results shown in Fig. \ref{fig:juli_paper}(b). These figures reveal that in B-doped systems, the FCC fraction decreases significantly with increasing concentration (dropping to approximately 71\% at 10 at.\%), accompanied by the emergence of an HCP phase. In contrast, the FCC structure remains almost fully intact in N-doped systems, even at 10 at.\%. This result quantitatively reproduces the findings of the original study \cite{DOLEZAL2025121221}, which reported distinct effects on structural stability depending on the type of light interstitial element, where B tends to segregate and distort the structure, whereas N tends to disperse and preserve it (Fig. \ref{fig:juli_paper}(c)). This confirms that the agent correctly implemented, executed, and analyzed the simulations. 
Furthermore, given that each simulation required approximately 16 - 43 hours and a total of 35 runs were performed, executing and supervising this workload purely by hand would have been extremely demanding, as it would require repeated job submission, continuous monitoring of job status to detect numerical failures, and resubmission of failed runs over multiple days. In practice, these calculations would not have been completed without \textit{PARC} autonomously handling these operations and maintaining the parallel execution of the runs.

However, not all tasks were completed perfectly, and our analysis revealed minor implementation errors. Specifically, the structural optimization did not account for cell shape relaxation. 
Furthermore, the agent did not implement the feature specified in the instruction for Trial Move 3, which said “attempt placement in the second nearest-neighbor shell if the first nearest-neighbor shell is fully occupied.” These errors went undetected in the self-assessment, possibly due to the complexity of the tasks.

On the other hand, we observed numerous instances where self-assessment and self-feedback worked effectively to prevent potential failures.
For example, the agent identified a flaw in the initial implementation where interatomic distances were computed assuming an orthogonal simulation cell, leading to incorrect nearest-neighbor relationships in tilted cells, and corrected this by adopting a minimum-image implementation to handle arbitrary cell shapes. In addition, it took measures to ensure the success of computationally expensive simulations by checking parameters before execution and adjusting them to more physically appropriate values.
Without these modifications, it is highly likely that errors would have been detected after the completion of simulations. This capability to autonomously maintain and rectify methodological validity enables the high stability of \textit{PARC} in long-horizon tasks.

\subsection{Case Study: Non-equilibrium MD Simulations under an Electric Field}
\label{sec:efield_md}

In this case study, we attempted to replicate the main results of a study \cite{hisama2023} on oxygen-ion conduction in yttria-stabilized zirconia (YSZ) under an external electric field. YSZ is a prototypical oxide-ion conductor used as an electrolyte in solid oxide fuel cells (SOFCs), where oxygen-ion transport under electric fields is a key factor in device performance and efficiency. This task required not only using existing tools but also extending a standard MD simulation package to handle external electric fields. 
Through this case study, we show that such a challenging task can result in failures that currently make fully autonomous use of \textit{PARC} difficult, but that the results it produces can be made useful when experts inspect and correct them. This case study also highlights remaining issues that need to be addressed in future work to achieve more reliable autonomous operation. 
 
\begin{figure}[H]
\centering
\includegraphics[width=0.95\linewidth]{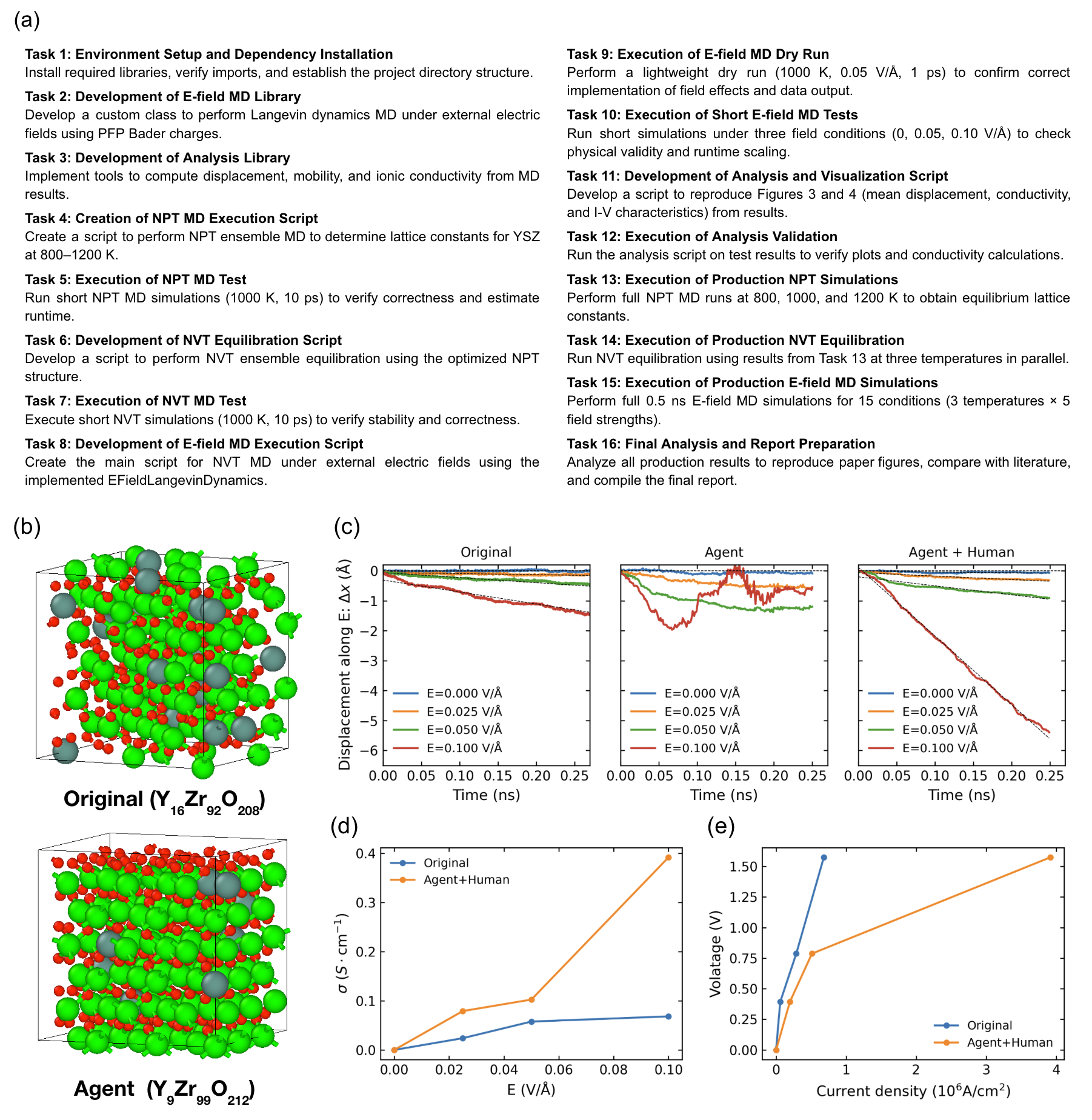}
\caption{\label{fig:hisama_paper}
Key outputs of \textit{PARC} on MD simulations of yttria-stabilized zirconia (YSZ) under an external electric field.
(a) Overview of the task sequence generated by the planner.
(b) YSZ structure generated by \textit{PARC}. While the original paper used a composition of $\mathrm{Y_{16}Zr_{92}O_{208}}$, \textit{PARC} generated an incorrect structure with lower Y content ($\mathrm{Y_{9}Zr_{99}O_{212}}$).
(c) Time evolution of oxygen atomic displacement at 800 K under various external electric field strengths. The ``Original'' shows the displacement re-plotted from data used in the original study (Ref.~\cite{hisama2023}); ``Agent'' shows the displacement analyzed by \textit{PARC}; and ``Agent + Human'' shows the displacement derived from human analysis of the agent's simulation trajectory using Eq.~(1).
(d) Ionic conductivity at each electric field strength (calculated based on Eq.~(2) of the original paper).
(e) Voltage dependence on current density (calculated based on Eqs.~(3) and (4) of the original paper).
}
\end{figure}

We provided \textit{PARC} with only the PDF file of the target paper and an initial structure file and instructed it to ``Create and execute code to reproduce the results of Figure 3 and Figure 4 (oxygen atom displacement, electric field dependence of ionic conductivity, etc.) in the paper.'' Regarding the initial structure, since it was difficult for the agent to generate the structure solely from the literature as evidenced by the failed structure shown in Fig.~\ref{fig:hisama_paper}(b), where while the original study used a composition of $\mathrm{Y_{16}Zr_{92}O_{208}}$, \textit{PARC} generated an incorrect structure with lower Y content ($\mathrm{Y_{9}Zr_{99}O_{212}}$), we instructed it to use the correct structure prepared by the authors in advance as the initial value.

Following this instruction, the planner devised a 16-step task sequence (Fig.~\ref{fig:hisama_paper}(a)). This plan constitutes a workflow for reproducing the paper, including the implementation of Langevin dynamics considering an external electric field, small-scale testing and parameter tuning, followed by simulation execution and analysis. Based on this plan, \textit{PARC} implemented an algorithm that applies an electric field-derived force ($F=qE$) in addition to interatomic interactions within an existing MD library, and then utilized it to execute simulations under various temperature and electric field conditions and analyzed the results.

Although \textit{PARC} encountered several issues detailed later, it managed to proceed as far as the simulation execution. However, during the preparation and analysis stages the agent failed in key steps. In the analysis code development, it failed to correctly implement the calculation of atomic displacements across periodic boundary conditions, which caused the overall task to end in failure. This failure led to the wrong results shown in Fig.~\ref{fig:hisama_paper}(c) - ``Agent". In addition, as mentioned below, \textit{PARC} failed to properly execute the NPT simulations for equilibration and failed to correctly implement and execute NPT simulations to determine the equilibrium lattice constant.

Since the implementation and simulation phases were completed, we analyzed the final simulation results ourselves. As shown in Fig.~\ref{fig:hisama_paper}(c) - ``Agent + Human", (d), and (e), we obtained data for the mean square displacement (MSD) of oxygen atoms, the electric field dependence of ionic conductivity, and current-voltage (I–V) characteristics. These results qualitatively reproduce the findings reported in the original study \cite{hisama2023}, suggesting that the simulation itself, as implemented and executed by the agent, was likely appropriate. However, because \textit{PARC} failed to properly execute the NPT simulations for equilibration, the structure used for the MD calculations under an external electric field did not perfectly match that of the original paper. The several-fold difference of the oxygen atom displacement observed in Fig.~\ref{fig:hisama_paper}(c) is possibly attributed to this discrepancy.

Despite these failures, multiple issues that arose in processes other than the analysis were successfully resolved by \textit{PARC}. For instance, when the agent failed to correctly implement and execute NPT simulations to determine the equilibrium lattice constant, which was an undesirable outcome in itself, it autonomously switched to an alternative method in which it performed energy minimization by scanning lattice constants to circumvent the problem. Furthermore, when simulations diverged under high-temperature conditions, the agent identified the cause as improper parameter settings such as the time step and successfully performed recalculations by adjusting them to appropriate values. These behaviors demonstrate the capability of \textit{PARC} to evaluate the validity of its approaches and apply corrections as necessary.

\subsection{Case Study: Kaggle Competition (NeurIPS Open Polymer Prediction Challenge 2025)}

In this case study, we applied \textit{PARC} to a model construction task targeting the Kaggle competition ``NeurIPS Open Polymer Prediction Challenge 2025" \cite{kaggle_polymer}. The objective of this competition is to create a model that predicts five physical properties from the SMILES representation of polymers: density, glass transition temperature ($T_g$), thermal conductivity ($T_c$), radius of gyration ($R_g$), and free fractional volume (FFV). To complete this task, the agent is required to execute a comprehensive data science workflow, which includes feature extraction from SMILES, selection of appropriate model architectures for each property, hyperparameter optimization, and model training.

\begin{figure}[H]
\centering
\includegraphics[width=0.5\linewidth]{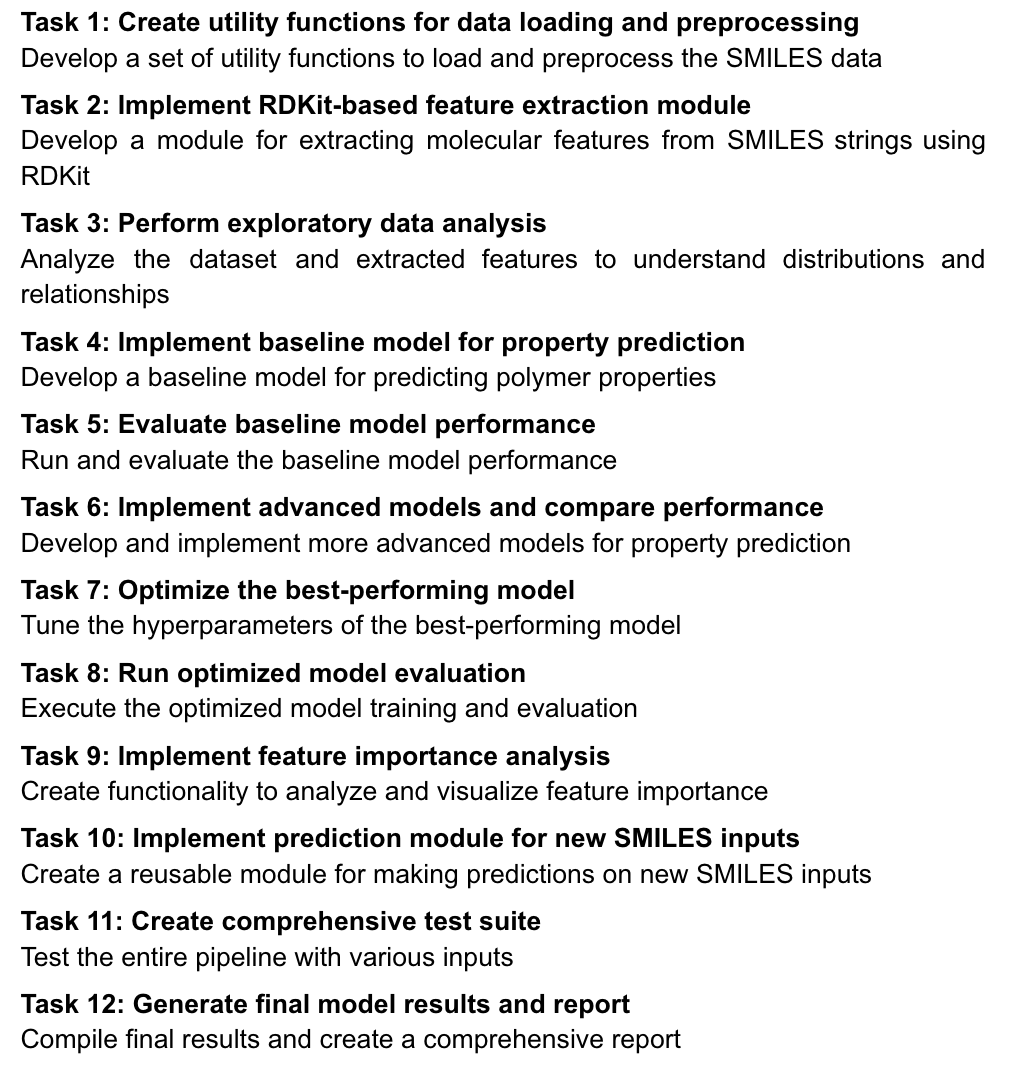}
\caption{\label{fig:kaggle_polymer}
Overview of the task sequence for the Kaggle competition ``NeurIPS Open Polymer Prediction Challenge 2025" generated by the planner.
}
\end{figure}

We provided the \textit{PARC}'s planner with the competition overview and dataset description as a text file and the dataset (CSV file), along with a very simple instruction: "create a model that can win this competition." Given that the use of the feature extraction tool "mordred" \cite{Moriwaki2018} is known to significantly impact the score in this competition, we also conducted an evaluation in which supplementary information suggesting the use of mordred was provided. Upon receiving these instructions, the planner formulated a task sequence consisting of 12 steps, as shown in Fig. \ref{fig:kaggle_polymer}. Since the generated plans remained largely consistent regardless of the presence of the supplementary information, we focus here on the plan generated with the supplementary information. This plan encompasses a comprehensive workflow, starting with basic statistical analysis of the data and proceeding to feature extraction, construction of a baseline model (Random Forest), implementation of models using LightGBM~\cite{lgbm} and XGBoost~\cite{xgboost}, hyperparameter optimization, verification of physicochemical validity, and finally, the development and testing of the final prediction module.

Throughout the execution of the task sequence, \textit{PARC} was able to successfully complete all tasks while autonomously resolving multiple technical issues by leveraging its self-feedback mechanism. For instance, in Task 4 and Task 5, data leakage occurred in the initial implementation where the model was trained using the entire dataset. However, through the self-feedback, \textit{PARC} identified this training discrepancy, modified the code to ensure strict separation between training and test data, and further quantitatively verified the validity of the correction. Furthermore, during the hyperparameter optimization in Task 7, the agent discovered that the optimized model for FFV exhibited performance degradation compared to the baseline prior to optimization. In response to this result, instead of uniformly applying the optimization results, the agent made an adaptive decision to utilize the model from Task 6 for FFV while adopting the optimized models for the other properties, thereby preventing any degradation in the overall score.

Table \ref{table:polymer_score} presents the evaluation results of the prediction accuracy of the model constructed by \textit{PARC}. As comparison targets, we used DeepEvolve \cite{deepevolve,deepevolve_repo}, a previous study that tackled the same task, and one of the best public notebooks available during the competition\cite{kaggle_polymer_nb_human}. 
For comparability with prior work, we evaluate these methods using the coefficient of determination ($R^2$) adopted in \cite{deepevolve,deepevolve_repo}, rather than the official metrics used in the competition. The scores for \textit{PARC} and the public notebook were calculated using cross-validation on the public data. In contrast, it should be noted that the score for DeepEvolve was calculated using a single-split test set. Given the small size of the dataset, this score may be subject to statistical fluctuation.

Since DeepEvolve does not employ mordred, we compare it with the results of \textit{PARC} without mordred. In this comparison, \textit{PARC} achieved an average $R^2$ score of 0.669, exceeding the 0.603 recorded by DeepEvolve. Furthermore, when compared to the public notebook created by a competition participant, \textit{PARC} with information on mordred recorded an average $R^2$ score of 0.781, which surpasses the human baseline of 0.764. It should be noted that these results are limited to a comparison with public code and do not imply that \textit{PARC} has surpassed the top solutions of the competition or human experts. Additionally, enabling the autonomous discovery of external tools suitable for problem solving remains a challenge for future work. Nevertheless, it is a noteworthy result that a general-purpose agent, not specifically designed for data science, was able to construct a practical model that exceeds a human baseline through autonomous trial and error, starting from only extremely simple instructions.

\begin{table}[H]
  \centering
  \caption{
Benchmark $R^{2}$ scores for predicting five polymer properties (FFV, $R_g$, $T_g$, $T_c$, Density). Scores for \textit{PARC} (with/without mordred) and the public notebook were evaluated by cross-validation on the competition’s public dataset. Scores for DeepEvolve are taken from the results in its repository \cite{deepevolve_repo}.
}
  \vspace{1em}
  \label{table:polymer_score}
  \begin{tabular}{l c c c c c}
    \toprule
    Target & \# of data & \makecell{DeepEvolve\\ \cite{deepevolve_repo}} & \makecell{\textit{PARC} \\ without mordred} & \makecell{\textit{PARC} \\ with mordred} & \makecell{Public notebook \\ \cite{kaggle_polymer_nb_human}} \\
    \midrule
    FFV      & 7030 & 0.262 & 0.691 & 0.825 & 0.825 \\
    $R_g$       &  614 & 0.730 & 0.483 & 0.730 & 0.688 \\
    $T_g$       &  511 & 0.483 & 0.575 & 0.674 & 0.654 \\
    $T_c$       &  737 & 0.788 & 0.769 & 0.788 & 0.777 \\
    Density  &  613 & 0.751 & 0.826 & 0.889 & 0.874 \\
    \midrule
    Average $R^2$ & -- & 0.603 & 0.669 & 0.781 & 0.764 \\
    \bottomrule
  \end{tabular}
\end{table}

\subsection{Case Study: Kaggle Competition (Santa 2023 - The Polytope Permutation Puzzle)}

In this case study, we employ the Kaggle competition ``Santa 2023 - The Polytope Permutation Puzzle" \cite{kaggle_santa} as a benchmark for advanced implementation tasks that require trial and error. The objective of this task is to discover a sequence of moves that restores polyhedral puzzles such as the Rubik's Cube from a scrambled initial state to a solved state while minimizing the score given as the total number of moves to solve all puzzles. This problem poses a significant challenge due to its extremely vast state space, making it difficult to even find a solution using simple search algorithms.

\begin{figure}[H]
\centering
\includegraphics[width=1.0\linewidth]{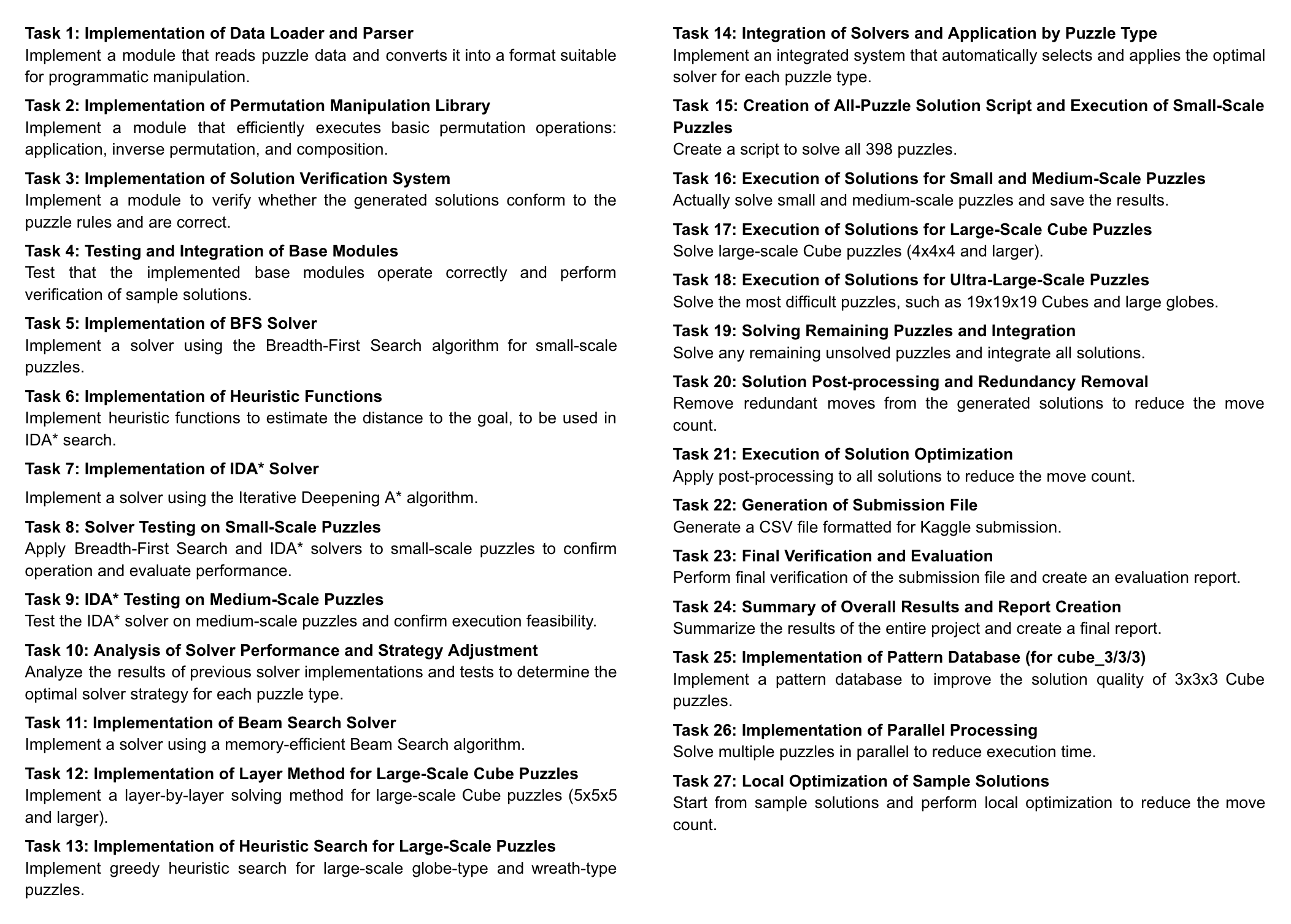}
\caption{\label{fig:kaggle_santa}
Overview of the task sequence for the Kaggle competition ``Santa 2023 - The Polytope Permutation Puzzle" generated by the planner.
}
\end{figure}

We provided \textit{PARC} with only the competition overview and dataset description as a text file and the puzzle data, accompanied by a minimal instruction: ``create a model that can win this competition." Upon receiving this instruction, the planner created a task sequence consisting of 27 steps, as shown in Fig.~\ref{fig:kaggle_santa}. This plan constitutes a comprehensive workflow, starting with the implementation of the puzzle environment (emulator) and proceeding to the verification of basic search algorithms, the adaptive selection of solution methods based on puzzle scale, and the post-optimization of solutions.

Throughout the execution phase, \textit{PARC} successfully completed all tasks while dynamically modifying its approach to address unexpected issues by leveraging its self-feedback. A representative example of this behavior is the strategic shift observed in Tasks 9 and 10. Initially, the agent implemented the ``Iterative Deepening A*" algorithm, but it detected a zero success rate and frequent timeouts during verification on medium-sized puzzles. In response to this result, the agent revised the plan and categorized all 398 puzzles into six groups based on size. After exploring a wide range of methods including simulated annealing and depth first search, the agent finally adopted beam search, which offered a superior balance between computational cost and success rate. 
Additionally, the agent utilized strategies suitable for the available computational resources, by calculating time complexity and testing on smaller systems, thereby avoiding problems of out-of-memory errors and computations running indefinitely.

Table~\ref{table:santa_score} presents the comparison between the results obtained by \textit{PARC} and the default solution defined as the inverse operations of the published shuffle sequence. Compared to this baseline, \textit{PARC} successfully reduced the total number of moves by approximately 21,000, improving the score from 1,220,590 to 1,199,430. A critical factor for score improvement is solving large-scale puzzles. However, \textit{PARC} was unable to solve these instances because it did not use existing external solvers, which were practically indispensable for top-tier solutions. This is one of the reasons why the result remains far from top solutions (the winning score is 53,770). Nevertheless, given that the score of one of the best public notebooks created by a human without using existing external solvers is approximately 1,160,000 \cite{kaggle_santa_zaburo_notebook}, the result can be considered sufficiently strong for a solution developed from scratch based on only simple instructions.

Overall, this case study demonstrates the effectiveness of \textit{PARC} for tackling complex software engineering tasks. The agent autonomously completed an engineering process that would typically require human engineers several days, including the implementation of a puzzle emulator, the selection of search algorithms, the implementation of parallel processing, and resource management.

\begin{table}[H]
  \centering
  \caption{
  Comparison of total moves (scores) achieved by \textit{PARC} and reference solutions on the Kaggle competition ``Santa 2023 - The Polytope Permutation Puzzle." Lower scores ($\downarrow$) indicate better performance.
  }
  \vspace{1em}
  \label{table:santa_score}
  \begin{tabular}{lcccc}
    \toprule

    & Default solution
    & \textit{PARC}
    & \makecell{High-scoring notebook\\ without external solvers \\ \cite{kaggle_santa_zaburo_notebook}}
    & Competition winner \\
    \midrule
    Total moves (Score)~$\downarrow$ & 1{,}220{,}590 & 1{,}199{,}430 & 1{,}158{,}978 & 53{,}770 \\
    \bottomrule
  \end{tabular}
\end{table}

\section{Conclusion}

As discussed in the introduction, we believe that the difficulty in long-horizon tasks stems not only from the capability of LLMs but also from the architectural constraints of existing agents. To overcome these limitations, we developed a new agent named \textit{PARC} that integrates planning, execution, self-assessment, and self-feedback into its architecture.

Through a series of case studies, we demonstrated that \textit{PARC} can autonomously and reliably execute complex tasks that conventional agents cannot solve without human intervention.
In the domains of data science and computational science, it successfully executed sequences of 10--20 tasks with roughly one hundred steps in total, which would require more than a day of work for a human. 
These results show that \textit{PARC} can autonomously perform trial-and-error to solve long-horizon tasks of a certain scale. They further show that our agent is not tailored to a single problem or domain but can serve as a general-purpose solution for long-horizon tasks.

However, our current approach and implementation do not eliminate all errors. In fact, we observed several errors that escaped detection with the self-assessment process during our experiments. To address these issues, we may be able to further improve the self-assessment and self-feedback to increase the pathways from error detection to correction. It would also be effective to optimize the structure and granularity of task decomposition by the planner. By breaking tasks down into units that are simpler and easier to verify, we can reduce the probability of errors occurring. Furthermore, improving the ability to autonomously discover and utilize external tools suitable for the task is crucial for solving complex problems.

If the behavior of standard coding agents is analogous to intuitive and immediate ``System 1" thinking, the self-assessment and self-feedback mechanisms introduced by \textit{PARC} can be interpreted as functioning as deliberative ``System 2" thinking. This study demonstrated that capabilities for long-horizon tasks can be substantially enhanced not by improving the underlying LLMs but by structuring thought and reasoning at the architectural level. The results achieved by \textit{PARC} suggest that more advanced AI can be realized not only through model scaling but also through the evolution of agent architectures. This represents an important step toward future agents that require autonomous trial-and-error, such as scientific discovery agents.
We anticipate that with the further development of architectures like ours, AI will soon evolve beyond mere efficiency tools to become capable of autonomously conducting tasks typically performed by human experts such as scientific discovery and complex large-scale software engineering.

\section*{Acknowledgments}
The authors are grateful to Dr. Kaoru Hisama for reviewing the results of the case study in Sec.~\ref{sec:efield_md} and providing valuable comments.

\printbibliography

\begin{appendices}
\section{Instruction used in the case study of Sec.~\ref{sec:lithium}}
\label{appx:lithium}

The following is the instruction actually provided to \textit{PARC} in the case study of Sec.~\ref{sec:lithium}.

\begin{lstlisting}
# Lithium Diffusion in Li10MP2S12-xOx Solid Electrolytes

## Objective
Calculate the activation energy of lithium ion diffusion from temperature dependence.

## Targets
- Basic Composition: Li10GeP2S12 (LGPS)
- Oxygen-Substituted System: Li10MP2S12-xOx (M = Ge; x = 0.5), Oxygen atoms replace sulfur atoms

### Crystal Structure Information
- Space group: P42/nmc (tetragonal)
- Initial structure: Use the experimentally determined crystal structure and lithium occupancy of LGPS (details described later)
- Oxygen-doped structure: Start from the most stable LGPS structure and consider all possible configurations where sulfur sites are substituted with oxygen

## Methods
### software
- Atomic Simulation Environment (ASE)
- Neural Network Potential: PreFerred Potential (PFP) version v3.0.0

### Calculation
Execute the following two types of simulations:
1. Determination of the most stable structure and calculation of lattice constants
2. Calculation of diffusion coefficients of Li via MD simulation.

## Tasks
### 1. Most Stable Structure and Lattice Constants
#### 1.1 Initial Structure
(Note): Refer to assets/element_pos.txt for element positions.
1.  The cell parameters and number of atoms for the initial structure are as follows:

```python
chemical_symbols = ["P"]*4 + ["Ge"]*2 + ["S"]*24 + ["Li"]*20
cell = [
    [8.7177, 0.0000,  0.0000],
    [0.0000, 8.7177,  0.0000],
    [0.0000, 0.0000, 12.634],
]
```

2. Generate 200 random Li configurations considering symmetry for the sites occupied by Li, and calculate the energy of each structure. The structure with the lowest energy is defined as the most stable structure.
Specifically, Li shall be placed at 16h, 8f, 4d, and 4c sites in the crystal.
For Li occupancy, treat these sites without distinguishing between symmetry-equivalent positions, and randomly select and place Li to satisfy the occupancy numbers for each site type (e.g., 8 in 16h, 6 in 8f, 4 in 4d, 2 in 4c).

3. Calculate the potential energy for the random structures mentioned above, and define the structure with the lowest energy as the most stable structure.

4. Based on the most stable structure above, generate oxygen-doped structures. For Li10GeP2S12-xOx (x = 0.5), substitute sulfur sites with oxygen, perform optimization of lattice constants and atomic positions for each structure, and calculate the energy. The structure with the lowest energy will be the initial structure for MD. Note that for the substitution of sulfur sites, consider all possible configurations among the 24 sulfur atoms. Both the positions and the unit cell need to be optimized.

### 2. Lithium Diffusion Calculation
Execute molecular dynamics simulations for the composition x=0.5 using the most stable structure obtained in step 1.

#### 2.1 Mean Squared Displacement (MSD) Plot
Horizontal axis: Time (ps)
Vertical axis: MSD (angstrom^2)
Plot: Time evolution at each temperature

MSD is defined as:
MSD(t) = (1/N) \sum_i <|ri(t) - ri(0)|^2>

Where N is the number of lithium atoms, and ri is the position of the i-th lithium atom.

#### 2.2 Arrhenius Plot
For (Li10MP2S12-xOx, M = Ge; x = 0.5):

Horizontal axis: 1000/T (1/K)
Vertical axis: log10(D) (cm^2/sec)

Calculate the diffusion coefficient D at each temperature by fitting from the linear region of the MSD.

#### 2.3 Calculation of Activation Energy
Calculate activation energy from the slope of the Arrhenius plot.

Important Notes:
- Self-determine all calculation parameters (cell size, simulation time, temperature range, time step, ensemble type, convergence criteria, etc.) required to generate the above outputs.
- Self-design the calculation procedures (dopant configuration generation method, selection criteria for the most stable structure, MD equilibration procedure, etc.).
- Self-evaluate the validity of output results based on qualitative and quantitative consistency with experimental values in papers or preceding studies.
\end{lstlisting}

\section{Instruction used in the case study of Sec.~\ref{sec:alloy}}
\label{appx:alloy}

The following is the instruction actually provided to \textit{PARC} in the case study of Sec.~\ref{sec:alloy}.

\begin{lstlisting}
## Goal
### Objective
Evaluate the segregation behavior and structural changes of light interstitial elements (B, N) in bulk Cr30Ni alloy (Cr: 30 at.%, Ni: 70 at.%).

### Strategy
Perform Monte Carlo (MC) calculations for the following 7 systems in total:
- Cr30Ni alloy
- Cr30Ni alloy doped with 1, 4, and 10 at.% B
- Cr30Ni alloy doped with 1, 4, and 10 at.% N

### Output
- Structures after MC execution.
- Plots of the structural abundance ratios of FCC, HCP, BCC, and Other analyzed using Polyhedral Template Matching (PTM) as a function of interstitial element concentration.

## Calculation Conditions
Please adhere to the following settings. Decide other parameters yourself.
- Simulation Cell:
    - Cr30Ni alloy: A 6x6x6 supercell of the FCC conventional cell, containing 259 Cr atoms and 605 Ni atoms.
    - The number of doped atoms for 1, 4, and 10 at.% is 9, 35, and 86, respectively.
- Interstitial Sites: Tetrahedral sites or octahedral sites.
- Potential: PFP v5.0.0 CRYSTAL_U0_PLUS_D3
- Convergence Condition for Structural Optimization: fmax=0.005 eV/A
- Number of MC Runs Required for Statistics: 5 runs for each composition.
- Number of MC Steps per Run: 6000.
- MC Trials: Choose from the following 5 options (Note: only option 1 is used for the undoped system):
    1. Swap atomic positions of metallic constituents: Select two metallic atoms of different chemical types and attempt to swap their positions within the lattice.
    2. Relocate a light interstitial atom near a new metallic host: Select a light interstitial atom and attempt to place it near a different metallic atom within its nearest neighbor shell.
    3. Introduce proximity between two light interstitial atoms: Select two light interstitial atoms and attempt to place one of them within the first nearest neighbor shell of the other. If the first shell is fully occupied, the placement is attempted in the second nearest neighbor shell.
    4. Separate two neighboring light interstitial atoms: If a pair of neighboring light interstitials exists, select one and attempt to move it away from its nearest interstitial neighbor. This move balances the proximity adjustment, ensuring an opportunity to promote both segregation and aggregation of light interstitials.
    5. Swap a metal neighbor of a light interstitial atom: For a selected light interstitial atom, identify one of its nearest metallic neighbors and swap its position with that of another metal atom of a different chemical type.
- MC Acceptance Criterion (Metropolis Method):
    - Temperature: 1073 K
    - Energy difference is calculated using the structure after structural optimization.
- Cutoff Radius for Nearest Neighbor Determination: 2.75 Angstrom
- PTM Analysis: Use PolyhedralTemplateMatchingModifier in OVITO with rmsd_cutoff=0, analyzing only the metallic lattice.
- Plot of PTM Analysis Results:
    - Create separate plots for B and N.
    - FCC (Blue), HCP (Orange), BCC (Green), Other (Red).
    - Display error bars.
\end{lstlisting}

\end{appendices}

\end{document}